\documentclass[10pt,journal,compsoc]{IEEEtran}
\usepackage{amsmath}
\usepackage{array}
\usepackage{amsthm}
\usepackage{mdwmath}
\usepackage{color}
\usepackage[justification=centering]{caption}
\ifCLASSOPTIONcompsoc
  \usepackage[nocompress]{cite}
\else
  \usepackage{cite}
\fi

\ifCLASSINFOpdf
   \usepackage[pdftex]{graphicx}
  \graphicspath{{../pdf/}{../jpeg/}}
  \DeclareGraphicsExtensions{.pdf,.jpeg,.png}
\else
  \usepackage[dvips]{graphicx}
  \graphicspath{{../eps/}}
  \DeclareGraphicsExtensions{.eps}
\fi

\begin{document}
\title{Image Disguise based on Generative Model}

\author{Xintao Duan, Haoxian Song, En Zhang and Jingjing Liu
\IEEEcompsocitemizethanks{\IEEEcompsocthanksitem Xintao Duan, Haoxian Song, En Zhang and Jingjing Liu were with School of Computer and Information Engineering, Henan Normal University, Xinxiang, Henan,
CHINA, 453007.\protect\\
E-mail: duanxintao@126.com
}
\thanks
}


\IEEEtitleabstractindextext{%
\begin{abstract}
To protect image contents, most existing encryption algorithms are designed to transform an original image into a texture-like or noise-like image, which is, however, an obvious visual sign indicating the presence of an encrypted image, results in a significantly large number of attacks. To solve this problem, in this paper, we propose a new image encryption method to generate a visually same image as the original one by sending a meaning-normal and independent image to a corresponding well-trained generative model to achieve the effect of disguising the original image. This image disguise method not only solves the problem of obvious visual implication, but also guarantees the security of the information.

\end{abstract}

}

\maketitle

\IEEEpeerreviewmaketitle

\IEEEraisesectionheading{\section{Introduction}\label{sec:introduction}}

\IEEEPARstart{C}{loud} has become a popular keyword in the computer society. Cloud computing technology provides individuals and organizations with a sufficiently large online space to store multimedia data, and offers people a convenient way to access and share data over the network. Due to the fact that these multimedia data may contain private, valued or even classified information, it is an important and urgent issue for individuals and organizations to prevent this important information from leakage. There are two common approaches: data hiding and encryption, to protect image contents from leakage. Data hiding technology embeds message into covers such as the image, audio or video, which not only protects the content of secret file, but also hides the communication process itself so that it cannot be attacked as far as possible \cite{Bao2015Image}. These methods have been proposed \cite{filler2010gibbs}, \cite{Holub2013Designing}, \cite{Holub2014Universal}. Data hiding has strong ability to resist detection, but can only achieve a relatively small payload, less than 1 bit per pixel. As the core of image security technology, image encryption is used to ensure the security of images. It uses a matrix feature of a digital image to change the position or value of a pixel in the spatial domain of the image, making the meaningful original image "messy ", lose its original appearance and transformed into information similar to the random noise of the channel. Although image encryption method is able to protect images with high security, the encrypted image formats are limited to noise and textures, making it easy to distinguish them from ordinary visually meaningful images. This feature similar to texture or noise is an obviously visible indication that the encrypted image may contain important information, it will draw people's attention, and cause a large number of attacks and analysis, thus increase the possibility of information leakage, loss and modification.

In order to solve the problems mentioned above, traditional method is used to combine two images together or divide the image into blocks \cite{Hou2016Image}, \cite{Hu2016An}. This paper presents a new image disguise method, we use $img$ and a well-developed model to generate $IMG'$, which is visually the same as the original image $IMG$, so when we deliver the original image $IMG$, we only need to pass $img$ and the model parameters to receiver without worrying about the risk of the original image being intercepted. As we know, this method has not been proposed so far.

\begin{figure}[ht]
\centering
\includegraphics[width=2.8in]{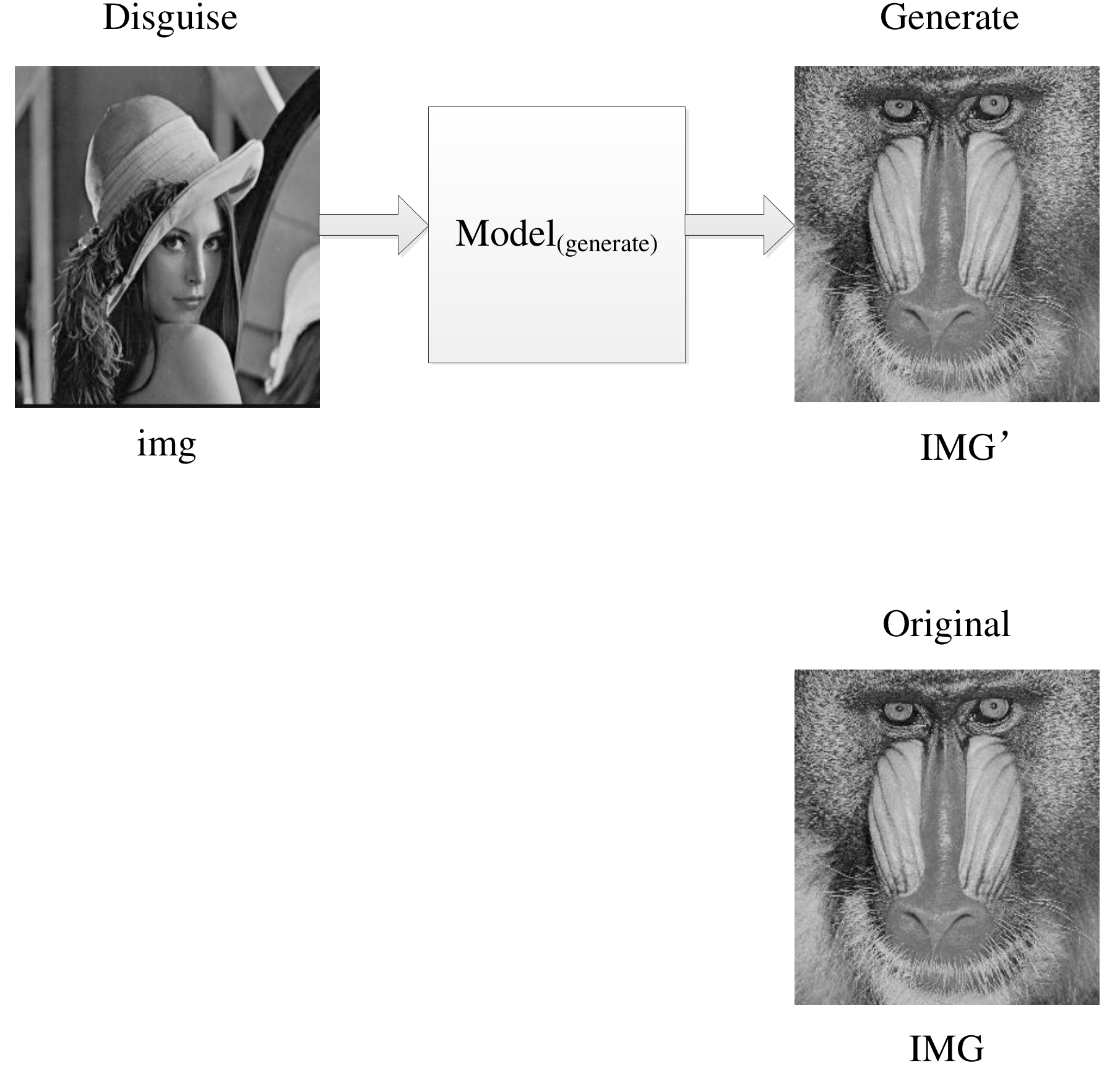}
\caption{}
\label{fig_1}
\end{figure}

As mentioned above, we proposed a new approach to disguise image, it can generate visually the same image as the original image by sending an image that is not related to the original image. The transmitted image is only a normal-meaningful image rather than the noise or texture encrypted image, and also achieve the effect the same as transferring the original image. This method avoids the attack caused by visual hint, and greatly improves the security of the image. To summarize, the major contributions of our work as below:

\begin{itemize}
\item{强调}Instead of transmitting the private, valuable or confidential image, we deliver an image $img$ that is irrelevant to the original image $IMG$, which drastically reduces the possibility of the original image being leaked.
\end{itemize}

\begin{itemize}
\item{强调}The image $img$ is not like texture or noise, and it does not give visual cues to attackers, which greatly reduces the possibility of being attacked.
\end{itemize}

\begin{itemize}
\item{强调}If the $img$ is intercepted, the attackers do not know it is a disguise image of the original image $IMG$. Even if the attackers know that, it takes a long time to verify these images one by one in the large data environment, which greatly extends the time of the image being attacked.
\end{itemize}

\section{Related work}
Restricted Boltzmann Machines (RBMs) \cite{Smolensky1986Information}, deep Boltzmann machines (DBMs) \cite{Salakhutdinov2009Deep} and their numerous variants are undirected graphical models with latent variables. The interactions within such models are represented as the product of unnormalized potential functions, normalized by a global summation/integration over all states of the random variables. This quantity and its gradient are intractable for all but the most trivial instances, although they can be estimated by Markov chain Monte Carlo (MCMC) methods. Mixing poses a significant problem for learning algorithms that rely on MCMC \cite{Bengio2012Better},\cite{Bengio2014Deep}. Deep belief networks (DBNs) \cite{Hinton2006A} are hybrid models containing a single undirected layer and several directed layers. While a fast approximate layer-wise training criterion exists, DBNs incur the computational difficulties associated with both undirected and directed models. Variational Auto-Encoders (VAEs) \cite{glorot2011deep} and Generative Adversarial Networks (GANs) \cite{Bengio2013Generalized} are well known to us. VAEs focus on the approximate likelihood of the examples, and they share the limitation of the standard models and need to fiddle with additional noise terms. Ian Goodfellow put forward GAN \cite{Goodfellow2014Generative} in 2014. Goodfellow theoretically proved the convergence of the algorithm, and when the model converges, the generated data has the same distribution as the real data. GAN provides a new training idea for many generative models and has hastened many subsequent works. GAN takes a random variable (it can be Gauss distribution, or uniform distribution between 0 and 1) to carry on inverse transformation sampling of the probability distribution through the parameterized probability generative model (it is usually parameterized by a neural network model), then a generative probability distribution is obtained. The GAN model includes a generative model G and a discriminative model D. The training objective of the discriminative model D is to maximize the accuracy of its own discriminator, and the training objective of generative model G is to minimize the discriminator accuracy of the discriminative model D. The objective function of GAN is a zero-sum game between D and G and also a minimum - maximization problem. GAN adopts a very direct way of alternate optimization, and it can be divided into two stages. In the first stage, the discriminative model D is fixed, the generative model G is optimized to minimize the accuracy of the discriminative model. In the second stage, the generative model G is the fixed in order to improve the accuracy of the discriminative model D. As a generative model, GAN is directly applied to modeling of the real data distribution, including generating images, videos, music and natural sentences, etc. Because of the mechanism of internal confrontation training, GAN can solve the problem of insufficient data in some traditional machine learning. Therefore, it can be used in semi-supervised learning, unsupervised learning, multi-view learning and multi-tasking learning. In addition, it has been successfully used in reinforcement learning to improve its learning efficiency. Although GAN is applied widely, there are some problems with GAN, difficulty in training, lack of diversity. Besides, generator and discriminator cannot indicate the training process. GANs offer much more flexibility in the definition of the objective function, including Jensen-Shannon, and all f-divergences \cite{Hinton2012Improving} as well as some exotic combinations. On the other hand, training GANs is well known for being delicate and unstable. The better discriminator is trained, the more serious gradient of the generator disappears, leading to gradient instability and insufficient diversity. WGAN (Wasserstein Generative Adversarial Networks \cite{Arjovsky2017Towards}, \cite{Arjovsky2017Wasserstein}) is an improvement to GAN, and it applies Wasserstein distance instead of JS divergence in the GAN. Compared to KL divergence and JS divergence, the advantages of Wasserstein distance are that it can still reflect their distance even if there is no overlap between the two distributions. At the same time, the problem of training stability and process indicating are solved.

Therefore, this paper chooses Wasserstein GAN so that guarantees training stability instead of GAN. It is no longer necessary to carefully balance the training extent between generator and discriminator. It basically solves the problem of collapse mode and ensures the diversity of samples.

\section{Approach}

\begin{figure}[ht]
\centering
\includegraphics[width=3.9in]{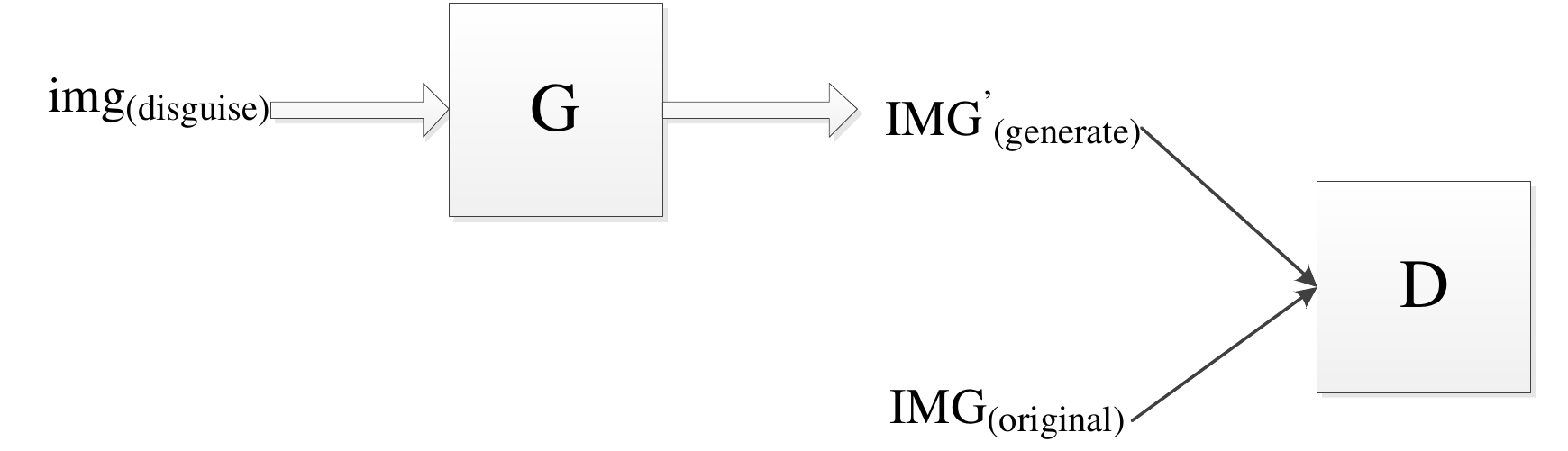}
\caption{}
\label{fig_2}
\end{figure}

\begin{figure}[ht]
\centering
\includegraphics[width=3.7in]{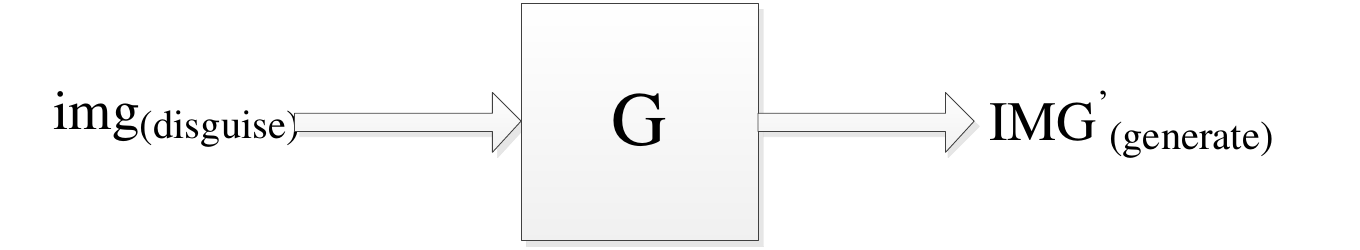}
\caption{}
\label{fig_3}
\end{figure}

The WGAN model is applied to generate the handwritten word by feeding the random noise $z$, but when the random noise $z$ is changed to a meaning-normal image $img$ which is independent of the original image, the model can still generate $IMG'$ visually the same as original image $IMG$. These several images taken from the standard set of images were evaluated in the paper, they are Lena, Baboon, Cameraman and Pepper, and they have the same size as 256 by 256. The purpose of this paper is to disguise an image using another image. The feed is a meaning-normal and independent image, and we train the generator through the WGAN, then it can generate the visually same as the original image with the disguise image and the trained generator. The flow charts of the whole experiment are shown in Fig.~\ref{fig_2} and Fig.~\ref{fig_3}


Wasserstein distance is also called the EM(Earth-Mover) distance
\[W({P_r},{P_g}) = \mathop {\inf }\limits_{\gamma  \sim \prod ({p_{r,}}{p_g})} {E_{(x,y) \sim \gamma }}[\parallel x - y\parallel ]\]

Where 
$\prod {({P_r},{P_g})} $ denotes the set of all joint distributions 
$\gamma (x,y)$ whose marginal are respectively ${P_r}$ and  ${P_g}$ . Intuitively, $\gamma (x,y)$  indicates how much “mass” must be transported from x to y in order to transform the distributions ${P_r}$   into the distribution ${P_g}$ . The EM distance then is the “cost” of the optimal transport plan.

The loss function of the generator in WGAN
\[ - {E_{x \sim {p_g}}}[{f_\omega }(x)]\]

The loss function of the discriminator in WGAN
\[{E_{x \sim {p_g}}}[{f_\omega }(x)] - {E_{x \sim {p_r}}}[{f_\omega }(x)]\]

\begin{figure}[ht]
\centering
\includegraphics[width=2.0in]{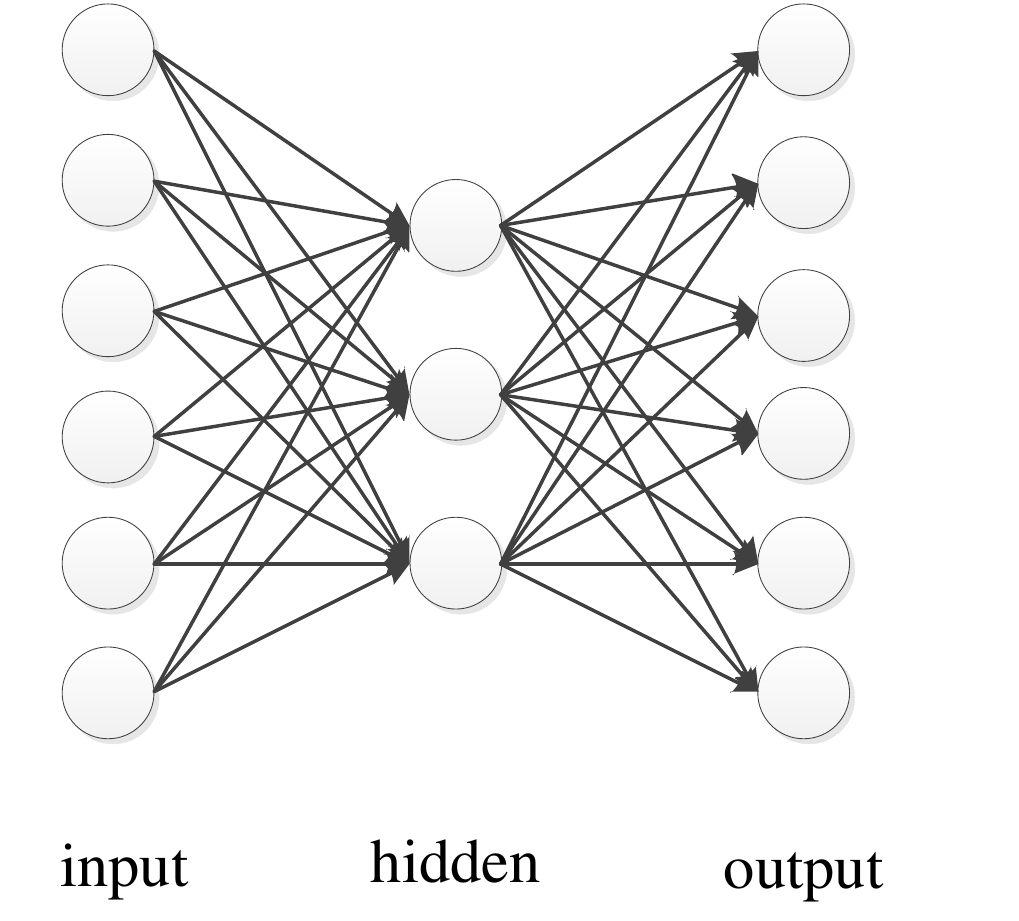}
\caption{}
\label{fig_4}
\end{figure}

In Fig.~\ref{fig_4}, the neurons number of the input layer, hidden layer, output layer is respectively 65536, 64 and 65536 in the generative model G.

In Fig. 5, the neurons number of the input layer, hidden layer, output layer is respectively 65536, 64 and 1 in the discriminator model D.

\begin{figure}[ht]
\centering
\includegraphics[width=2.0in]{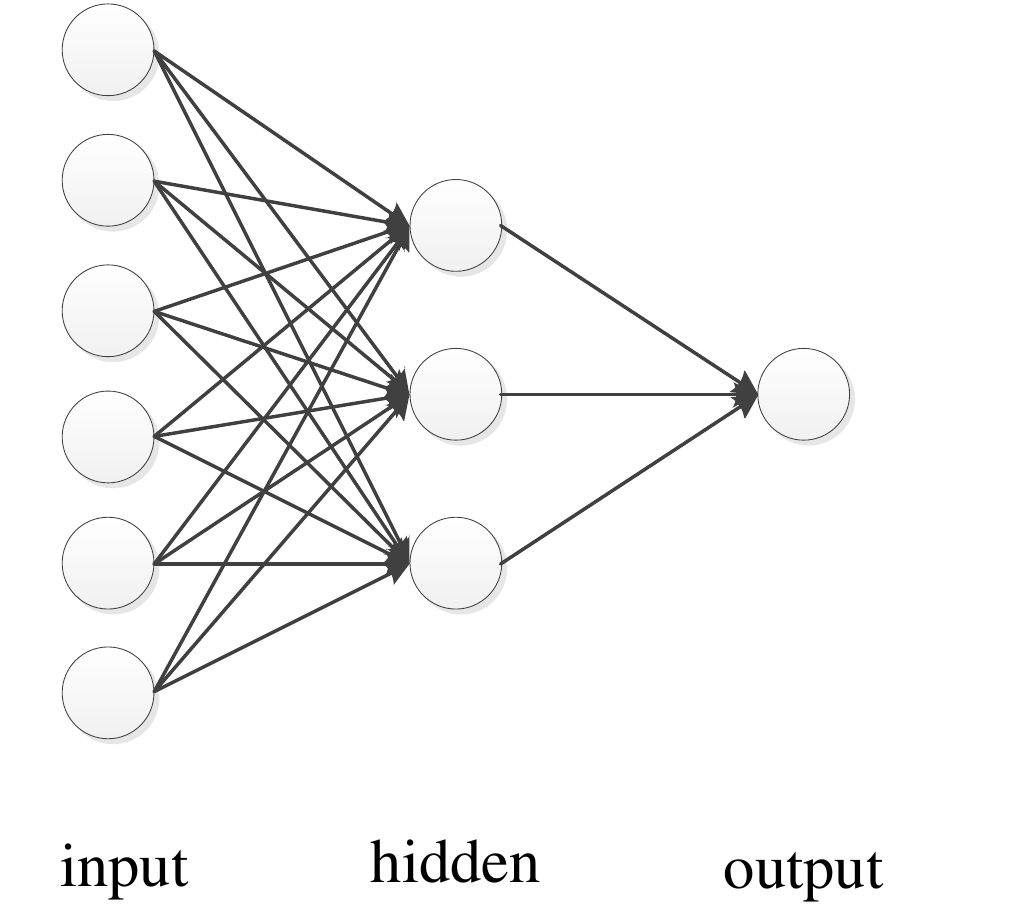}
\caption{}
\label{fig_5}
\end{figure}
\section{Experiments}

\begin{figure}[ht]
\centering
\includegraphics[width=2.6in]{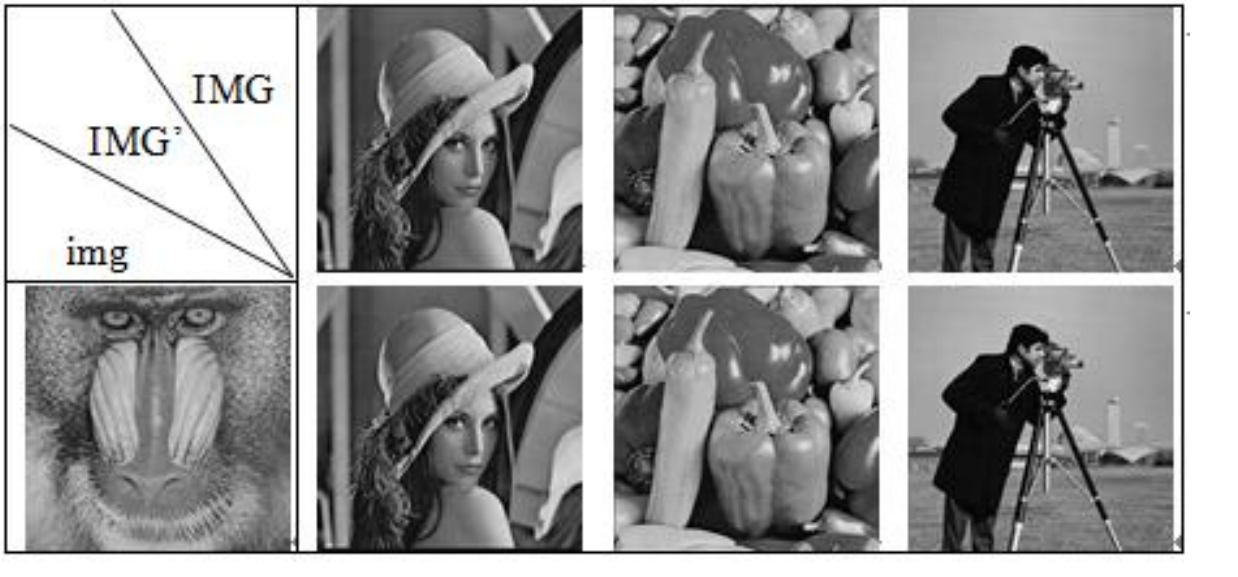}
\caption{}
\label{fig_6}
\end{figure}

\begin{figure}[ht]
\centering
\includegraphics[width=2.6in]{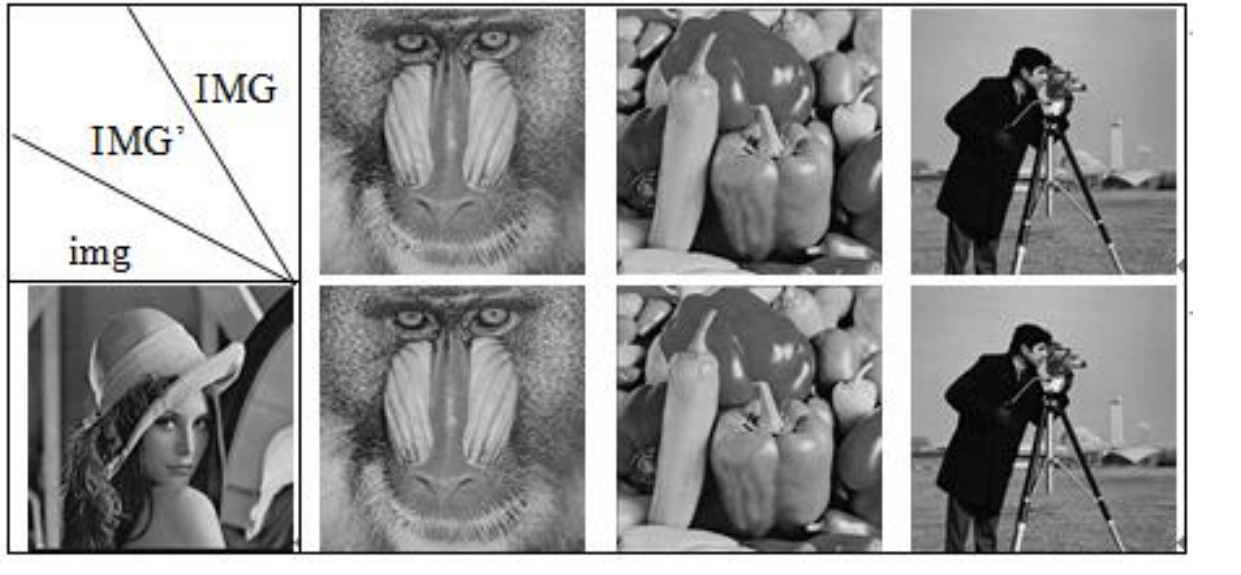}
\caption{}
\label{fig_7}
\end{figure}

In this paper, 5,000 images are randomly selected from the CelebA dataset to experiment, and the result shown that the image disguise method can be implemented well. As shown in Fig.~\ref{fig_6} and Fig.~\ref{fig_7}, we feed a meaningful image $img$ into the generative model, generating the $IMG'$ that is the visually same as the original image $IMG$.

As shown above, we choose Baboon as the disguise image $img$, it can generate different $IMG'$ through corresponding well-trained generative model. It can be seen that these original images $IMG$ can be entirely represented by those images $IMG'$.

We change the disguise image Baboon in Fig.~\ref{fig_6} to Lena. Using the same method, Lena can also generate the $IMG'$, visually the same as these original images $IMG$. Similarly, we take the Peppers, Cameraman to experiment respectively, and get the identical result. We save corresponding the generative model G1, G2 and G3 of the disguise image Baboon, Peppers and Cameraman generating the $IMG'$ Lena respectively, and apply them to the next experiment, instead of the WGAN.

\begin{figure}[ht]
\setlength{\abovecaptionskip}{0.cm}
\centering
\includegraphics[width=2.5in]{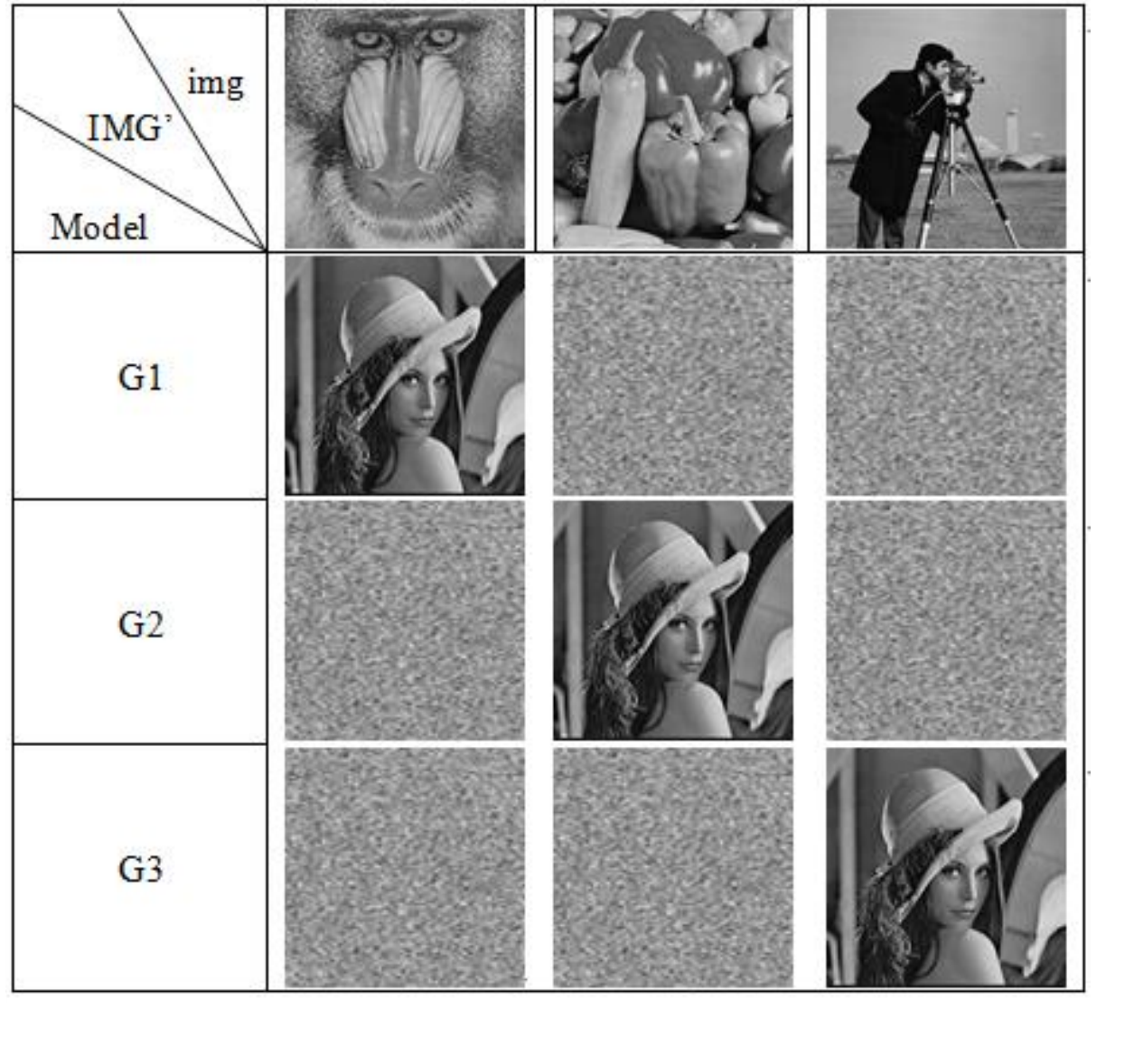}
\caption{}
\label{fig_8}
\end{figure}

As shown in Fig.~\ref{fig_8}, we feed the img Baboon, Peppers and Cameraman to the G1, G2 and G3 respectively, we found that Baboon can generate Lena only by G1, Peppers can generate Lena only by G2, and Cameraman can generate Lena only by G3. We save the corresponding generative model G4, G5 and G6 of the disguise image Lena, Peppers and Cameraman generating the $IMG'$ Baboon to experiment respectively.

\begin{figure}[ht]
\setlength{\abovecaptionskip}{0.cm}
\centering
\includegraphics[width=2.6in]{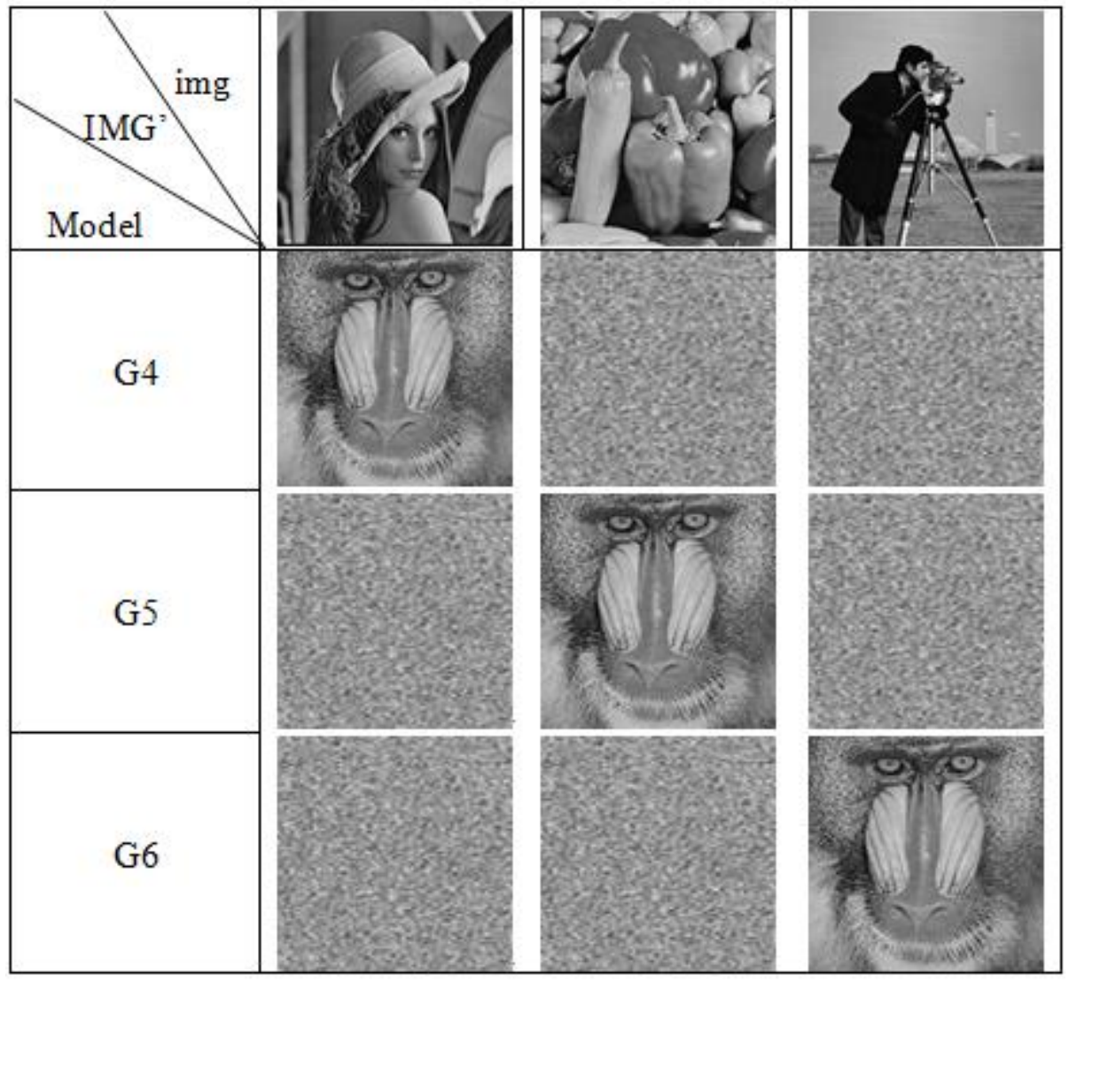}
\caption{}
\label{fig_9}
\end{figure}

As shown in Fig.~\ref{fig_9}, the disguise images are Lena, Peppers and Cameraman respectively, Baboon is given as the original image. The result shows that Lena can generate Baboon only by G4, Peppers can generate Baboon only by G5, and Cameraman can generate Baboon only by G6.

In this experiment, we have successfully achieved the effect of image disguise by feeding a normal-meaningful image to generate an image visually the same as the original image, and when the original image is given, the generative model corresponding disguise image is unique and specific. Consequently, the image disguise method proposed in this paper is feasible. In practical application, we are more concerned with the content of the image rather than the pixels in addition to professional image workers, this method can produce an image visually the same as an original image, which can satisfy most requirements, thereby, we suppose that if you want to send an image, you only need to feed a meaningful image to the corresponding generative model, generate an image visually the same as the original one, no needing direct transmission of the original image. This method avoids leakage of the private image information, and improves the security of the image.

\section{Conclusion}

To sum up, the paper proposed a new image disguise method. An image visually the same as the original image is generated by feeding a normal-meaningful image to the generative model. A fed image corresponds uniquely to a generative model. This method is practical. Therefore, it can be applied to image disguise and image protection.

\section*{Acknowledgment}

This work was supported by the National Natural Science Foundation of China under Grant NO.U1204606, U1404603, the Science and Technology Foundation of Henan Province under Grant No.172102210335.

\ifCLASSOPTIONcaptionsoff
  \newpage
\fi

\bibliographystyle{IEEEtran}
\bibliography{ref}

%

%







\end{document}